\documentclass[letterpaper]{article} 
\usepackage{aaai23}  
\usepackage{times}  
\usepackage{helvet}  
\usepackage{courier}  
\usepackage[hyphens]{url}  
\usepackage{graphicx} 
\usepackage{natbib}  
\usepackage{caption} 
\usepackage{algorithm,algpseudocode}
\usepackage{amsmath}
\usepackage{amsfonts}
\usepackage{amssymb}

\usepackage{newfloat}
\usepackage{listings}
\usepackage{amsmath}
\usepackage{subcaption}
\usepackage{booktabs}
\usepackage{multirow}

\DeclareCaptionStyle{ruled}{labelfont=normalfont,labelsep=colon,strut=off} 
\floatstyle{ruled}
\newfloat{listing}{tb}{lst}{}
\floatname{listing}{Listing}
\title{Mahalanobis-Aware Training for Out-of-Distribution Detection}
\author{
    Connor Mclaughlin\thanks{Work done at MIT LL; now at Northeastern University}, Jason Matterer\thanks{Work done at MIT LL; now at STR}, and Michael Yee
}
\affiliations{
     MIT Lincoln Laboratory\\
    mclaughlin.co@northeastern.edu, jason.matterer@str.us, myee@ll.mit.edu
}

\nocopyright

\begin{document}

\maketitle

\begin{abstract}
While deep learning models have seen widespread success in controlled environments, there are still barriers to their adoption in open-world settings. One critical task for safe deployment is the detection of anomalous or out-of-distribution samples that may require human intervention. In this work, we present a novel loss function and recipe for training networks with improved density-based out-of-distribution sensitivity. We demonstrate the effectiveness of our method on CIFAR-10, notably reducing the false-positive rate of the relative Mahalanobis distance method on far-OOD tasks by over 50\%.
\end{abstract}

\section{Introduction}
Deep neural networks have achieved widespread ubiquity across decision-making processes in various applications, ranging from medical imaging to autonomous driving. When deployed to the open world, these systems frequently encounter samples from classes not represented in the training data, known as ``out-of-distribution" (OOD) data. In these settings, it is essential for the model to have the ability to abstain or hand off the difficult decision to a human expert. Consequently, the development of effective OOD detection techniques is crucial for enhancing the robustness and reliability of these systems in real-world scenarios.

Existing OOD detection methods typically fall under two categories: (1) model output-based methods, which take into account the predicted logits or probabilities of the model as an indicator of confidence, or (2) model representation-based methods, which measure the similarity of intermediate layer representations to those seen at training time. Our study focuses on the latter class of methods, which rely on learning a representative model of the in-distribution (ID) data in order to compute the likelihood of test samples. Some studies assume a Gaussian structure for the data \cite{lee2018simple}, achieving empirical success but leaving much of the theory unexplored. In contrast, others adopt more complex, non-parametric methods \cite{sun2022out} to avoid imposing any assumptions on the data. In this paper, we aim to contribute to this discourse by investigating the following question: Can we improve the performance of Gaussian-based OOD detection methods through explicit model training aimed at creating Gaussian-like data representations? 

Our contributions are twofold: (1) we first present a novel regularization loss that better aligns the training-time objective and test-time OOD detector, and (2) we also provide a training recipe with noise reduction techniques that make our method more accessible to a limited computational budget. Our proposed method results in  a significant leap in sensitivity to OOD instances, all while maintaining minimal disruption to in-distribution performance. 
\begin{table}[tp!]
\centering
\begin{tabular}{@{}ccccc@{}}
\toprule
\multirow{2}{*}{Method} & \multicolumn{2}{c}{Far-OOD} & \multicolumn{2}{c}{Near-OOD} \\ \cmidrule(l){2-5} 
                        & FPR95 $\downarrow$        & AUROC $\uparrow$        & FPR95 $\downarrow$        & AUROC $\uparrow$       \\ \midrule
MSP                    &  50.91            & 90.62             & 61.25              & 89.67             \\
Energy                 &  40.97            &  89.10            & 51.30            &  86.24            \\
KNN                     &  37.10            &  94.47           & 52.31              &  89.67            \\
ViM                     &  27.50            &  95.66            & 55.25              &  87.66            \\
RMD                     &  36.58            &  93.80            & 51.54              &  89.01            \\ \midrule
Ours                    &  \textbf{17.79}            &  \textbf{96.76}            & \textbf{49.53}              &  \textbf{90.43}            \\ \bottomrule

\end{tabular}
\caption{Results using CIFAR-10 as in-distribution dataset.}
\label{cifar10table}
\end{table}



\section{Methods}
The task of OOD detection hinges on learning a scoring function $S(\mathbf{x})$ which captures the similarity of test data to the training distribution. In the case of density estimation methods, this scoring function is akin to the likelihood function of a probabilistic model representing the in-distribution data. Combined with a threshold $\tau$, the OOD decision rule can be formalized as follows:

\[
\text{Decision}(\mathbf{x}) =
\begin{cases}
\text{ID}, & \text{if } S(\mathbf{x}) \geq \tau \\
\text{OOD}, & \text{if } S(\mathbf{x}) < \tau
\end{cases}
\]

The Mahalanobis-Distance method \cite{lee2018simple} models the distribution of latent representations of the neural network as coming from a class-conditional Gaussian with means $\boldsymbol{\mu}_{1..k}$ and a tied covariance matrix $\boldsymbol{\Sigma}$. The most commonly selected representation is the output of the penultimate layer of the network, denoted as $\mathbf{z} = F(\mathbf{x})$ for input $\mathbf{x}$. The scoring function is given as the Mahalanobis distance of representation $\mathbf{z}$ to the closest in-distribution class centroid (with flipped sign so ID samples have higher score):

\[
\text{MD}_k(\mathbf{z}) = (\mathbf{z} - \boldsymbol{\mu}_k)^T \boldsymbol{\Sigma}^{-1} (\mathbf{z} - \boldsymbol{\mu}_k)
\]

\[
S_{\text{MD}}(\mathbf{z}) = -\underset{k}{\text{min}{\hspace{2pt}} } \text{MD}_{k}(\mathbf{z})
\]

As the success of Mahalanobis distance relies upon ID samples having higher likelihoods than OOD samples, our key insight is in promoting the likelihood of ID data throughout training. Our proposed training objective utilizes an online estimate of Gaussian parameters in order to compute the predicted probability of test samples under Bayes' rule:
\begin{align*}
    P(Y=k|X=x) &= \frac{P(Y=k)P(X=x|Y=k)}{\sum_{{k'}}P(Y=k')P(X=x|Y=k')}
    \\
    &= \frac{\exp(\mathrm{MD}_k)}{\exp(\sum_{{k'}}\mathrm{MD}_{k'})} 
\end{align*}

We then look to minimize the cross-entropy loss using these predictions. Our final combined loss is a weighted combination of the initial cross-entropy loss using model output logits ($L_{\mathrm{base}}$) and the cross-entropy loss using Mahalanobis distances as logits ($L_{\mathrm{maha}}$). We introduce a hyperparameter $\alpha$ to control the balance between losses:
\begin{align*}
    L_{\mathrm{reg}} = (1-\alpha)L_{\mathrm{base}} + \alpha{}L_{\mathrm{maha}}
\end{align*}

The final component of our proposed method is in estimating the Gaussian parameters throughout training. As the batch size $n$ may be much smaller than the dimensionality of our feature representations $d$, we introduce two necessary components to reduce the noise introduced by small batch estimates.  First, we use a shrinkage estimator for the covariance matrix \cite{ledoit2004well} rather than the maximum likelihood estimator. Second, we maintain a moving average (EMA) of the means and covariance which is updated with each batch of training data. 

Once the training is complete, our method utilizes the Relative Mahalanobis (RMD) score \cite{ren2021simple} variation of Mahalanobis distance as the OOD scoring function. 

\section{Results}
\begin{figure}[h]
    \centering
    \includegraphics[width=\linewidth]{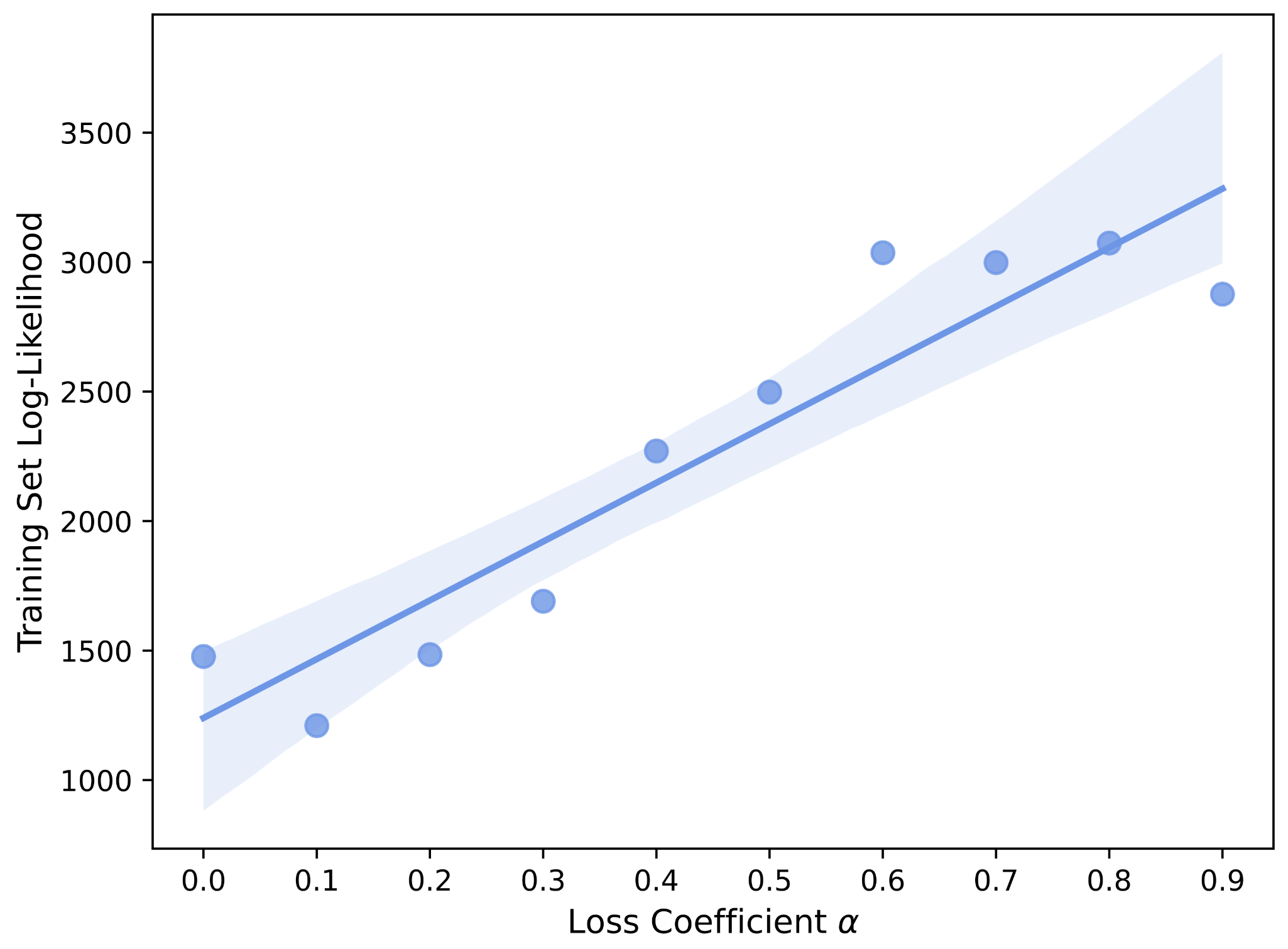}
    \caption{Training Dataset Gaussian LL vs. Loss $\alpha$}
    \label{fig:gaussianality}
\end{figure}

\begin{figure}[h]
    \centering
    \includegraphics[width=\linewidth]{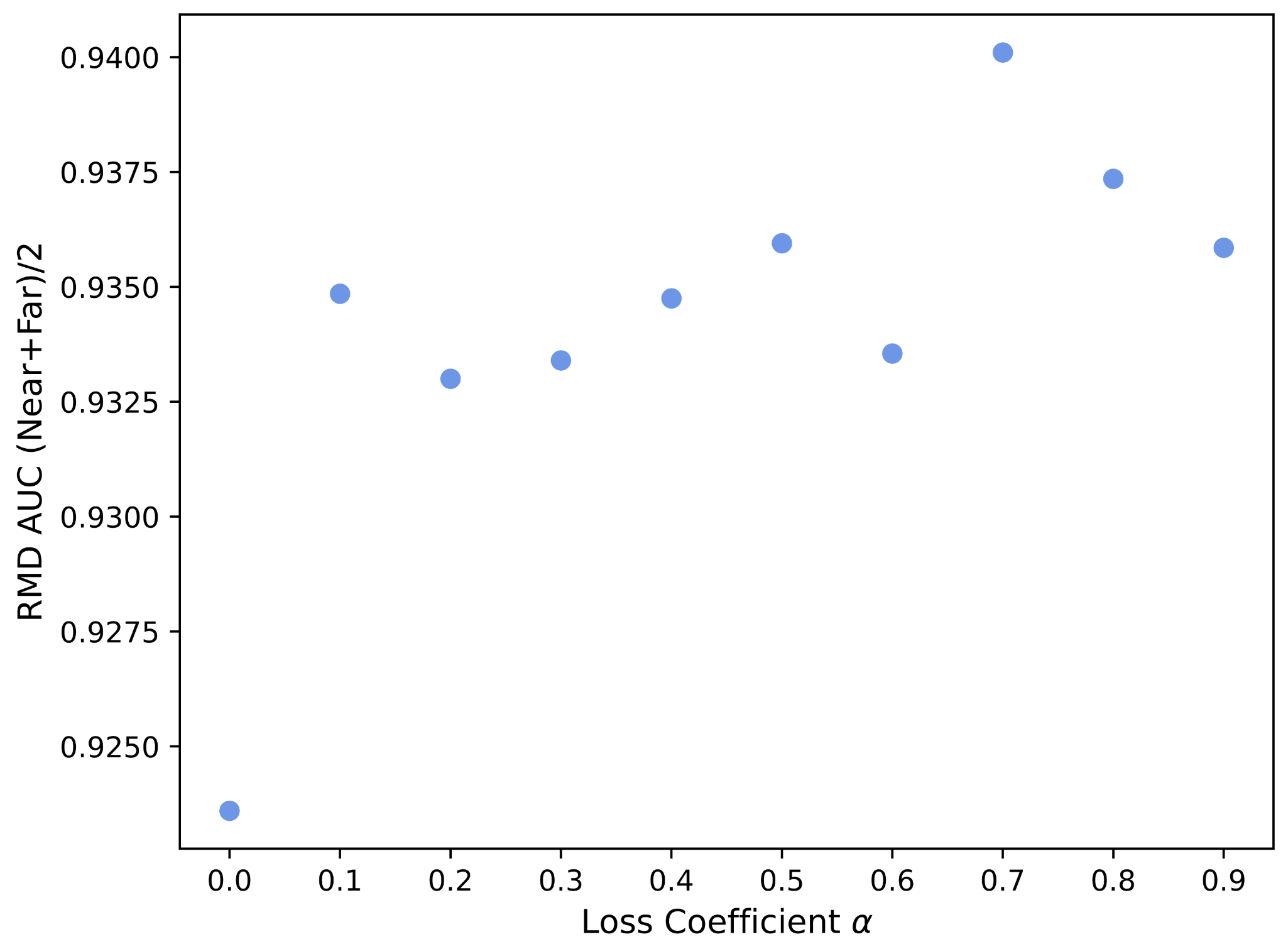}
    \caption{Sensitivity Analysis of Parameter $\alpha$}
    \label{fig:sensitivity}
\end{figure}

We demonstrate the efficacy of our training policy through a series of experiments using CIFAR-10 as the in-distribution dataset. We use the ResNet18 architecture and follow the baseline CIFAR-10 setup provided by OpenOOD \cite{yang2022openood}. Table \ref{cifar10table} compares our method to recent approaches on a far-OOD benchmark consisting of (SVHN, Places365, iSUN, LSUN, and Textures), and a more challenging near-OOD benchmark consisting of CIFAR-100. We compare our method to MSP \cite{hendrycks2016baseline}, Energy \cite{liu2020energy}, RMD \cite{ren2021simple}, KNN \cite{sun2022out}, and ViM \cite{wang2022vim}. Our method outperforms existing baselines in both near-OOD and far-OOD settings, making it a reliable choice regardless of the deployment environment. We additionally show that our Mahalanobis loss has the intended effect on the Gaussian likelihood of data representations in Figure \ref{fig:gaussianality}. Our method is not sensitive to the added hyperparameter $\alpha$, and consistently outperforms the baseline model as shown in Figure \ref{fig:sensitivity}. In-distribution performance remains steady, with our method achieving 94.7\% accuracy compared to the 95.2\% accuracy of the baseline recipe.

\section{Conclusion}
In summary, we present a novel training recipe based on Mahalanobis distance regularization for improved out-of-distribution detection. Future work includes exploring the scalability of this method to large scale datasets.

\section{Acknowledgements}
DISTRIBUTION STATEMENT A. Approved for public release. Distribution is unlimited.

This material is based upon work supported by the Under Secretary of Defense for Research and Engineering under Air Force Contract No. FA8702-15-D-0001. Any opinions, findings, conclusions or recommendations expressed in this material are those of the author(s) and do not necessarily reflect the views of the Under Secretary of Defense for Research and Engineering.

\bibliography{aaai23}

\end{document}